%% file: IEEE-conference-template-062824.tex
\documentclass[conference]{IEEEtran}
\IEEEoverridecommandlockouts

\usepackage[T1]{fontenc}
\usepackage{mathptmx}  

\usepackage{cite}
\usepackage{amsmath}
\usepackage{amssymb}
\usepackage{amsfonts}
\usepackage{algorithm}
\usepackage{algpseudocode}
\usepackage{graphicx}
\usepackage[english]{babel}
\usepackage{textcomp}
\usepackage{xcolor}
\usepackage{booktabs}
\usepackage{expl3}
\usepackage{longtable}
\usepackage{adjustbox}
\usepackage{hyperref}
\usepackage{orcidlink}
\usepackage{array}

\begin{document}

\title{Curriculum Learning in Genetic Programming Guided Local Search for Large-scale Vehicle Routing Problems\\
}
\author{
\IEEEauthorblockN{
Saining Liu\,\orcidlink{0000-0001-5667-3325}, 
Yi Mei\,\orcidlink{0000-0003-0682-1363}, 
Mengjie Zhang\,\orcidlink{0000-0003-4463-9538}
}
\IEEEauthorblockA{
\textit{ Centre for Data Science and Artificial Intelligence \& School of Engineering and Computer Science}\\
\textit{Victoria University of Wellington, PO Box 600, Wellington 6140, New Zealand} \\
\{Saining.Liu, Yi.Mei, Mengjie.Zhang\}@ecs.vuw.ac.nz
}
}
\IEEEoverridecommandlockouts
\IEEEpubid{\makebox[\columnwidth]{ 979-8-3315-3431-8/25/\$31.00~\copyright2025 IEEE \hfill} 
\hspace{\columnsep}\makebox[\columnwidth]{ }}
\IEEEpubidadjcol
\maketitle

\begin{abstract}
Manually designing (meta-)heuristics for the Vehicle Routing Problem (VRP) is a challenging task that requires significant domain expertise. 
Recently, data-driven approaches have emerged as a promising solution, automatically learning heuristics that perform well on training instances and generalize to unseen test cases. 
Such an approach learns (meta-)heuristics that can perform well on the training instances, expecting it to generalize well on the unseen test instances.
A recent method, named GPGLS, uses Genetic Programming (GP) to learn the utility function in Guided Local Search (GLS) and solved large scale VRP effectively.
However, the selection of appropriate training instances during the learning process remains an open question, with most existing studies including GPGLS relying on random instance selection.
To address this, we propose a novel method, CL-GPGLS, which integrates Curriculum Learning (CL) into GPGLS. Our approach leverages a predefined curriculum to introduce training instances progressively, starting with simpler tasks and gradually increasing complexity, enabling the model to better adapt and optimize for large-scale VRP (LSVRP).
Extensive experiments verify the effectiveness of CL-GPGLS, demonstrating significant performance improvements over three baseline methods.
\end{abstract}

\begin{IEEEkeywords}
curriculum learning, genetic programming guided local search, large-scale vehicle routing problem, utility function.
\end{IEEEkeywords}

\section{Introduction}
The Vehicle Routing Problem (VRP) is a classical combinatorial optimization problem \cite{dantzig1959truck} that aims to determine optimal routes for a fleet of vehicles to service geographically distributed customers while minimizing operational costs and adhering to constraints such as vehicle capacity or time windows \cite{laporte2007you}. 
Large-Scale Vehicle Routing Problems (LSVRP) is a special case of the Capacitated VRP when the number of customers is more than 200 \cite{arnold2019efficiently, gendreau2010solving, sabar2015math, costa2020adaptive}.
The manual design of (meta-)heuristics for the VRP is a complex and knowledge-intensive task. 
Effective (meta-)heuristics often require significant domain expertise to craft problem-specific components, which can be both time-consuming and error-prone \cite{blum2003metaheuristics}. 
As the problem scale increases, designing heuristics capable of efficiently navigating the vast solution spaces becomes increasingly difficult.
Consequently, developing effective, domain-independent heuristics for large-scale problems remains a significant challenge in optimization research.

Recently, data-driven approaches have emerged as a compelling alternative for automating the process of designing effective (meta-)heuristics. These approaches utilize machine learning models to automatically discover patterns in training data and generate heuristics that perform effectively across diverse problem instances. 
For example, Genetic Programming (GP) and data-mining have been applied to evolve rules for routing problems, demonstrating notable success in solving vehicle routing problems \cite{costa2020adaptive, wang2021survey}. 
A core principle of these methods is that the learned heuristics, trained on representative problem instances, can generalize to unseen test cases, addressing the need for scalability and adaptability in complex optimization scenarios \cite{bengio2009learning, laporte2009fifty}. 
However, a critical question remains: how to effectively select training instances during the learning process. 

Random instance selection, a common approach in the training of meta-heuristics, often fails to consider the inherent structure and characteristics of problem instances, leading to inefficient learning, slower convergence, and suboptimal generalization to unseen instances \cite{jin2018data}. To overcome these limitations, we introduce curriculum learning (CL) into the training process. Inspired by human learning, CL organizes training instances in a structured sequence based on difficulty, enabling models to build foundational knowledge with simpler instances before tackling more complex ones \cite{soviany2022curriculum}. This progressive training strategy enhances learning efficiency, accelerates convergence, and reduces the likelihood of getting trapped in local optima, ultimately improving solution quality.

The overall goal of this paper is to enhance the training process of GPGLS by incorporating the CL strategy, thereby improving its effectiveness in solving LSVRP. 
GPGLS, which integrates the flexibility of GP with the robust local search capabilities of Guided Local Search (GLS), has demonstrated its effectiveness in addressing complex optimization problems \cite{arnold2019efficiently}.
The objectives of this work can be summarized as follows:
\begin{itemize}
    \item Develop a CL strategy tailored for training GPGLS, employing fixed transitions between difficulty levels after a number of training generations to structure the learning process effectively.
    \item Validate the proposed CL strategy through its integration into GPGLS, demonstrating significant improvements in solving LSVRP than other three methods, which use different training strategy.
\end{itemize}

The remainder of this paper is organized as follows: The proposed method is presented in Section \ref{CL-GPGLS}. 
Section \ref{Experiments} presents the experimental design, preliminary investigation and results analysis. Finally, Section \ref{conclusion} concludes the paper and outlines directions for future research.

\begin{figure}[!t]
\centerline{\includegraphics[width=0.45\textwidth]{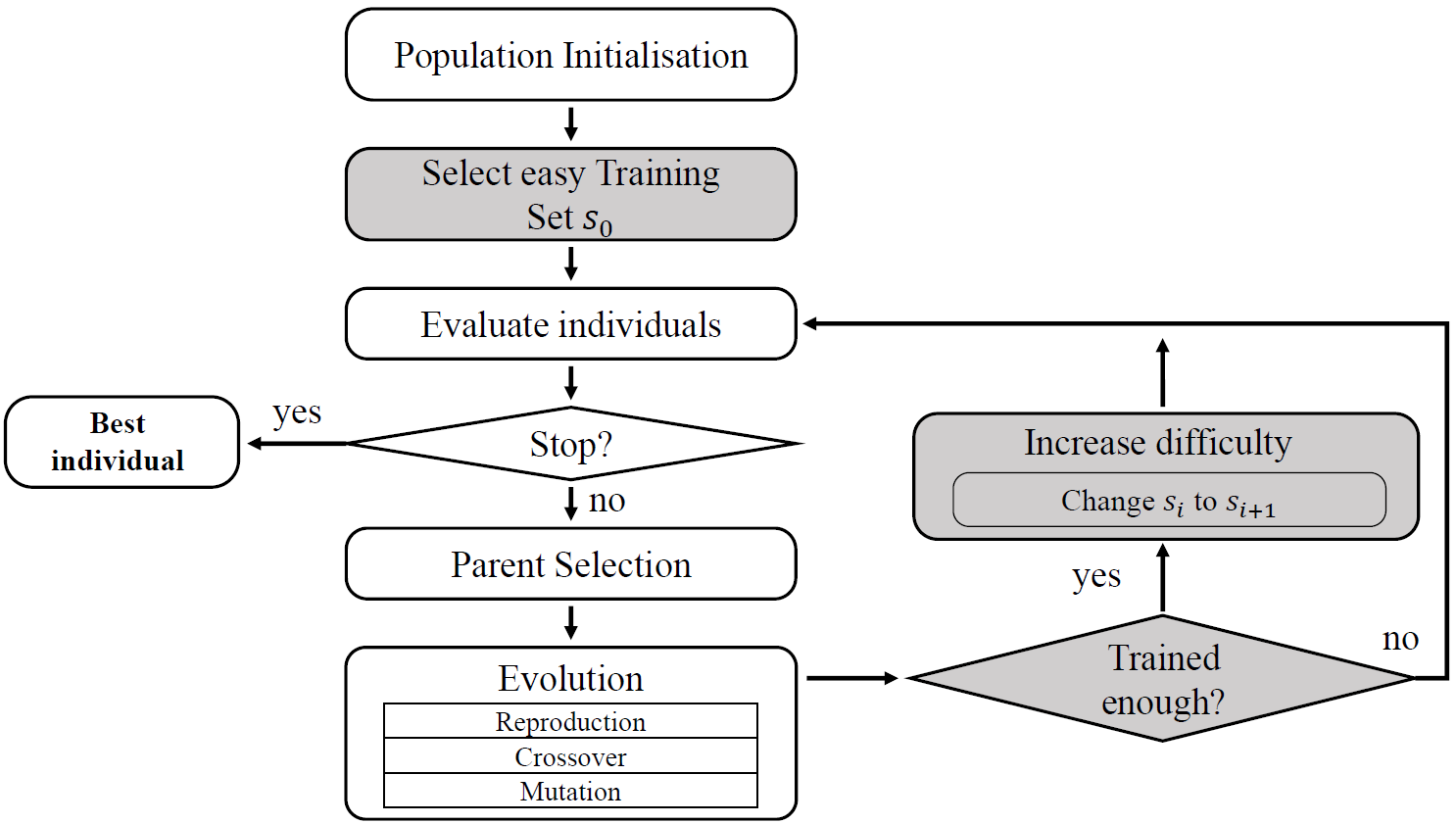}}
\caption{The evolutionary framework of CLGPGLS. The colored steps are the newly proposed steps compared with the original GPGLS.}
\label{flowchart}
\end{figure}

\section{CL-GPGLS} \label{CL-GPGLS}

\subsection{Overall Framework}
The GPGLS follows an offline learning process for GLS, with its primary innovation being the automatic design of utility functions for GLS that consider a broader range of factors, including instances, solutions, and edges \cite{liu2024gpgls}. Each individual in the population corresponds to a potential utility function for GLS. The framework consists of two main components: the GP component and the GLS evaluator. The GP component is responsible for generating and evolving a population of GP trees, each representing a candidate utility function, while the GLS evaluator processes the GP population alongside a training set. In this process, each GP individual serves as the utility function within GLS, guiding the search for features to penalize. The fitness of each individual is determined by the quality of the solutions produced by GLS when applied to the training set.

The uncolored part of Fig. \ref{flowchart} depicts the GP portion of the GPGLS framework, while the colored sections highlight the novel contributions of this paper. Initially, the GP population is generated using the ramp-half-and-half method. The population is then evaluated to determine the fitness of each individual. Subsequently, GP iteratively evolves the population based on these fitness values, applying genetic operators such as crossover and mutation to generate new individuals. This iterative process continues until the stopping criterion is met. Meanwhile, the GLS evaluator assesses the individuals by applying them to improve the solutions for VRP instances.

To enhance the capability of GPGLS in solving LSVRP, a CL strategy is incorporated into the training phase of the model. 
As illustrated in the colored sections of Fig. \ref{flowchart}, a training set selection mechanism is introduced during both the initialization and evolution phases of GP. 
The difficulty of the training set increases progressively throughout the training process, aligning with the core principles of CL.
More details can be found in section \ref{cl}

\begin{figure}[!t]
\centerline{\includegraphics[width=0.45\textwidth]{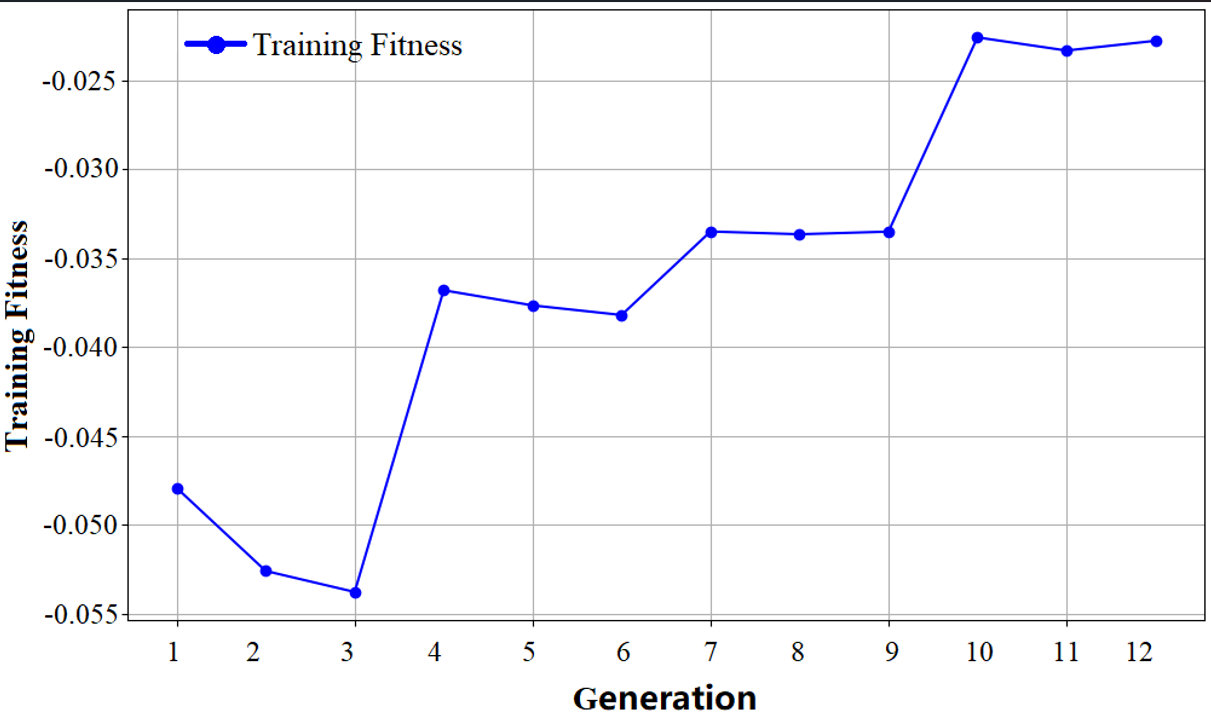}}
\caption{An example of the training process using the method proposed in this paper. The four steps correspond to four levels of difficulty in the training set.}
\label{train}
\end{figure}

\subsection{Curriculum Scheduler} \label{cl}

The difficulty of VRP instances increases with scale, leading to an exponentially growing solution space and more complex constraints. To address this, we implement a CL strategy that organizes training instances in ascending order of difficulty. 
The model begins with small-scale instances and progressively moves to larger, more complex ones, ensuring it learns to tackle increasingly difficult tasks. 
This gradual approach is effective for large-scale problems, allowing the model to incrementally build its capacity. 

In the flowchart of Fig. \ref{flowchart}, we highlight the key differences from traditional simplified algorithms by using darker shading. The initial population is first trained on an easy training set. At each generation, we check whether the current training set is trained enough. In other words, we check whether the model has met the transition criteria. If not, we continue training on the current set. If yes, we transition to a more challenging training set with higher difficulty.

Fig. \ref{train} illustrates an example of the training process. The horizontal axis represents the number of generations, while the vertical axis depicts the average fitness of the population. The fitness value for each generation is computed using the following formula:
\begin{equation}
fitness = \frac{final\_{cost} - init\_{cost}}{init\_{cost}},
\end{equation}
where \( final\_{cost} \) and \( init\_{cost} \) denote the solution cost after and before optimization, respectively. A lower fitness value indicates better performance.
As shown by the red dashed lines, the training process is shown as four distinct steps, corresponding to different levels of training sets. 
With each step, the fitness value decreases, signaling that the population is learning knowledge from the current training set of how to solve LSVRP. However, as the model encounters more challenging instances in later steps, the overall fitness value increase due to the increasing complexity of training set, suggesting that the model is still adapting to the new challenges.

The transition between training steps occurs before the population fully converges, driven by two key reasons: (1) knowledge from simpler sets may not generalize well to more complex ones, and (2) early transitions conserve computational resources, allowing more focus on progressively harder tasks. This strategy prevents overfitting to simpler instances while maintaining sufficient exploration at each stage to identify effective strategies. It fosters a balance between exploring different scales and refining the model’s understanding, ultimately discovering robust utility functions that can handle various problem instances. 

In summary, we have designed a curriculum scheduler for GPGLS to address the LSVRP.
From the flowchart, it can be seen that we haven't made big changes to the original framework, only adding a more refined training strategy. Often, small changes can lead to significant improvements, and we will present detailed experimental results in section \ref{results}. 
The four distinct steps in Fig. \ref{train} demonstrate that the training sets designed in this study progressively increase in difficulty, as reflected by the rising fitness values. 
If the test fitness results show progressive improvement, it would suggest that the knowledge gained from smaller-scale problems is beneficial for solving larger-scale ones.

\begin{table}[!t]
\centering
\footnotesize
\caption{Parameters for CL-GPGLS}
\begin{tabular}{cl}
\hline
\textbf{Parameter} & \textbf{Value} \\ \hline
Population size & 100 \\ 
Crossover probability & 0.8 \\ 
Mutation probability & 0.15 \\ 
Elite probability & 0.05 \\ 
Tournament size & 3 \\
Initial minimum/maximum depth & 2 / 6 \\ 
Initialization method & Ramped-half-and-half \\ 
Instances increase sequence & \{100, 300, 600, 1000\} \\
Evaluate time per instance & 5s * (scale/100) \\
Trained for per scale & 3 generations \\ 
 \hline
\end{tabular}
\label{tab:exp_params}
\end{table}

\section{Experimental Studies} \label{Experiments}

\subsection{Experimental Settings}
The datasets used in this study are generated by the generator provided by \cite{queiroga202110}, which follows a similar generation scheme of dataset X \cite{uchoa2017new}.
The four key parameters are configured as follows: random depot positioning, random customer positioning, unitary demand distribution, and medium average route size. The training set comprises three instances for each problem scale, and the testing set consists of 30 instances, each with a size of 1000.

To evaluate the effectiveness of CL, we conducted four groups of experiments. The first, ST (stochastic sequence), utilized randomly selected training sets for each generation. The second, RG (random grouped sequence), trained with a fixed scale of training sets for three generations before switching to a different scale. The third, LO (large-only sequence), used large-scale training sets that matched the test set size. Finally, the fourth group, CL (curriculum learning sequence), followed the proposed method, where training sets were arranged progressively from simple to complex tasks.

The terminal and function sets for GPGLS are identical to those used in previous work, comprising features extracted from instances, solutions, and edges. The function set \{+, -, *, ÷\} takes two arguments, where “÷” is protected and returns 1 when division by zero occurs. The parameter settings for GLS are consistent with those used in KGLS \cite{arnold2019knowledge}. 
Table \ref{tab:exp_params} outlines the parameters related to GP and CL. The training process incrementally increases the instance scale, with evaluation time adjusted proportionally to the instance size. At each scale, the model is trained on a fixed set of 3 instances for 3 generations, ensuring steady learning progression across different scales.

\subsection{Results and Analysis} \label{results}

\begin{figure}[!t]
\centerline{\includegraphics[width=0.45\textwidth]{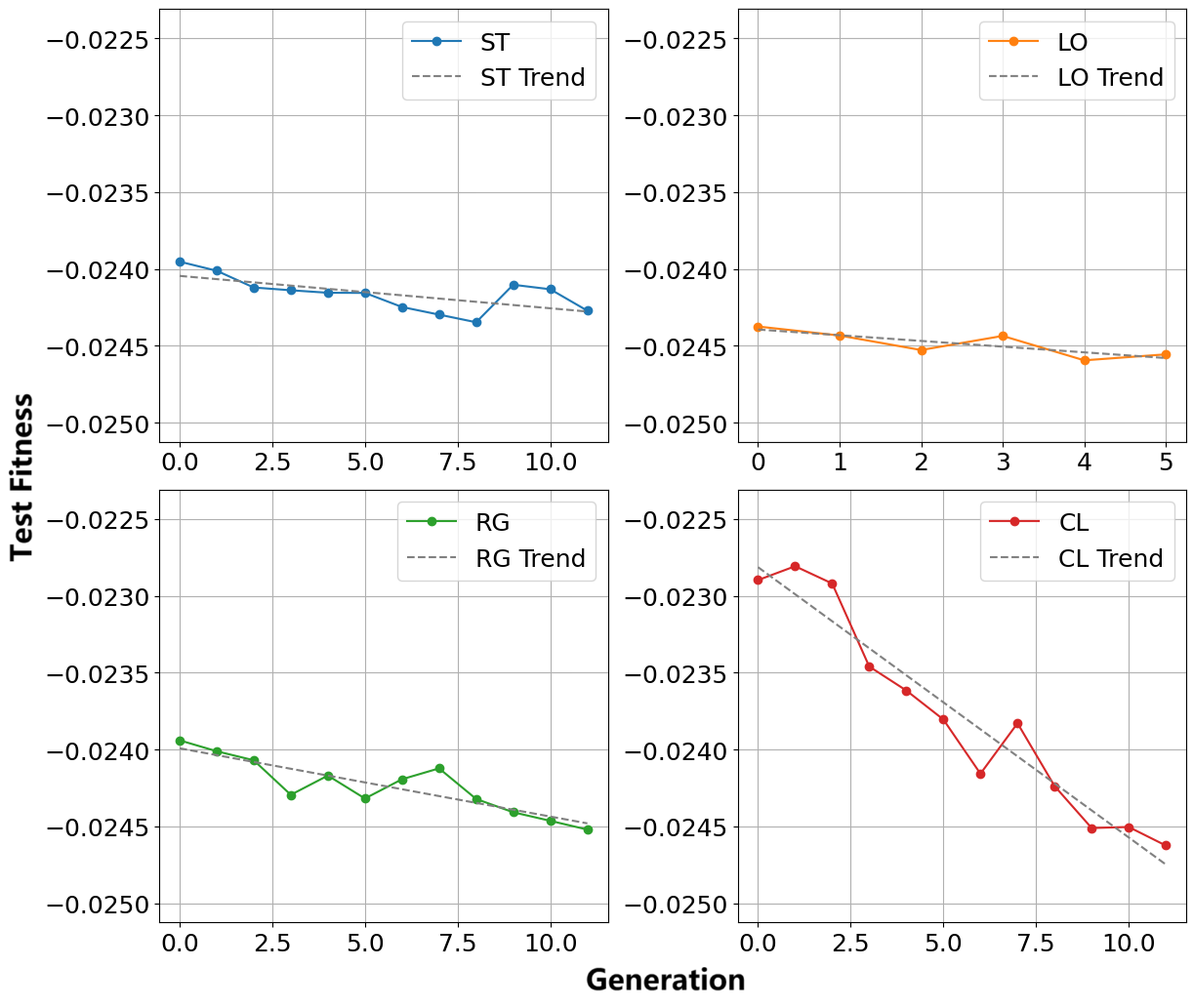}}
\caption{Test the fitness of four methods: ST (stochastic sequence), LO (large-only sequence), RG (random grouped sequence), and CL (curriculum learning sequence). The trend line shows that the method proposed in this paper achieves higher learning efficiency and outperforms the other three methods in the final generation.}
\label{test}
\end{figure}

Each experiment was conducted over 30 independent trials. 
To compare the performance of CL-GPGLS against the other methods, the Wilcoxon rank-sum test was applied at a significance level of 0.05. 
The statistical results show that CL-GPGLS is significantly better than ST-GPGLS in 14 of the 30 instances, 2 are significantly better than LO-GPGLS, and 3 are significantly better than RP-GPGLS. These three methods are not significantly better than CL-GPGLS in any instance.
Fig. \ref{test} presents the test fitness results for the four algorithms. It is clear that ST and LO exhibit relatively low learning efficiency, while RG shows slightly better efficiency but with occasional fluctuations. In contrast, CL demonstrates the highest learning efficiency, achieving superior fitness values in the final generation compared to the other algorithms. Although the knowledge gained from simpler instances does not perform as well on the test set, it expands the search space for more challenging instances. As a result, with the same training time, CL-GPGLS achieves better results by effectively leveraging this expanded search space.

\begin{figure}[!t]
\centerline{\includegraphics[width=0.47\textwidth]{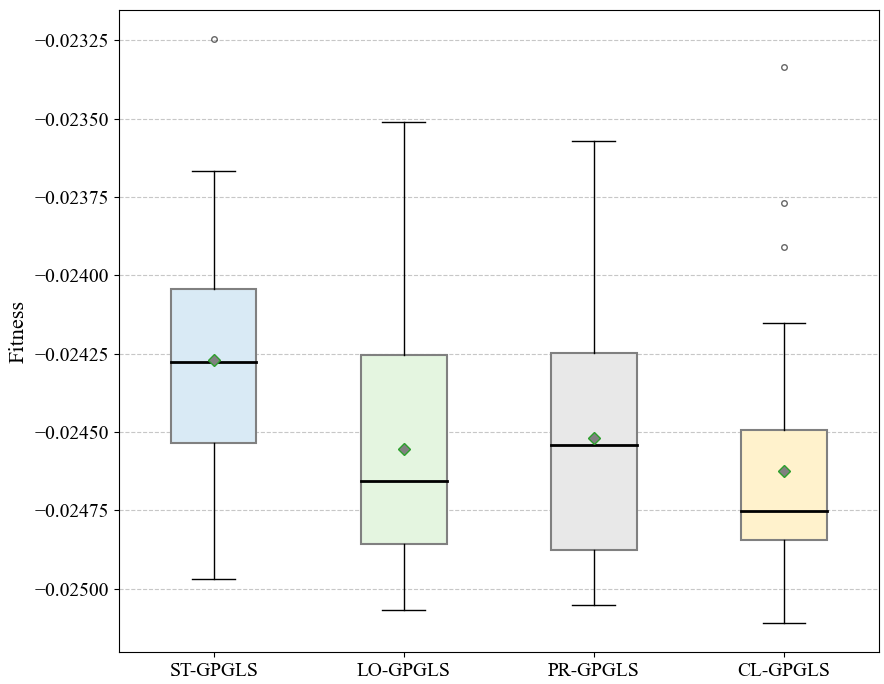}}
\caption{The box plots of the test performance of 30 independent runs of the proposed four methods. }
\label{last}
\end{figure}

Fig. \ref{last} visualizes the box plots of the test performance of 30 independent runs of the proposed CL-GPGLS and other three methods, where lower fitness values indicate better performance. 
CL-GPGLS performs best, with the smallest box, lowest median, and shortest whiskers, reflecting consistent optimization. However, the presence of three outliers suggests occasional instability in some test instances. 
ST-GPGLS, with a larger box and higher median, exhibits poorer performance and greater variability, as well as one outlier, indicating less consistent optimization.
LO-GPGLS and PR-GPGLS have similar box sizes, but LO-GPGLS has a lower median and the longest whisker, indicating more variability and less stable performance across instances.

In summary, CL-GPGLS demonstrates superior performance by providing lower mean costs and smaller variances in most cases, indicating greater stability than the competing methods. While there are some instances, such as instance 15, where ST-GPGLS performs slightly better in terms of mean cost, CL-GPGLS consistently excels in both cost and stability. These results emphasize the robustness and superior performance of CL-GPGLS compared to existing algorithms.

\section{Conclusions} \label{conclusion}
The goal of this paper is to utilize CL to improve the training process of GPGLS for solving LSVRP. This goal has been successfully achieved through the design of CL-GPGLS. Specifically, four training sets with varying levels of difficulty were employed, where the model is trained progressively, starting with simpler instances and advancing to more complex ones, thereby enhancing performance for large-scale problems.

The results demonstrate that the utility functions evolved by CL-GPGLS significantly outperform those derived from other training strategies. Furthermore, the training process exhibits four distinct steps across the training sets, confirming the successful design of the curriculum scheduler.

Some interesting directions for future work include the
design of more flexible curriculum schedulers, taking into
account factors such as convergence speed, training time, and
fitness deviation when changing training sets. 
In terms of improve performance of GLS, methods for adaptively selecting algorithm components, like search operators \cite{pei2025adaptive}, also own great potential for solving LSVRP \cite{pei2022investigation}.
The relationship between VRP instance difficulty and algorithm performance should also be further explored. Lastly, generating more
VRP datasets based on real-world scenarios could benefit researchers.


\input{IEEE-conference-template-062824.bbl}

\end{document}

%% file: IEEE-conference-template-062824.bbl

%% file: IEEE-conference-template-062824.bbl
\begin{thebibliography}{10}
\providecommand{\url}[1]{#1}
\csname url@samestyle\endcsname
\providecommand{\newblock}{\relax}
\providecommand{\bibinfo}[2]{#2}
\providecommand{\BIBentrySTDinterwordspacing}{\spaceskip=0pt\relax}
\providecommand{\BIBentryALTinterwordstretchfactor}{4}
\providecommand{\BIBentryALTinterwordspacing}{\spaceskip=\fontdimen2\font plus
\BIBentryALTinterwordstretchfactor\fontdimen3\font minus \fontdimen4\font\relax}
\providecommand{\BIBforeignlanguage}[2]{{%
\expandafter\ifx\csname l@#1\endcsname\relax
\typeout{** WARNING: IEEEtran.bst: No hyphenation pattern has been}%
\typeout{** loaded for the language `#1'. Using the pattern for}%
\typeout{** the default language instead.}%
\else
\language=\csname l@#1\endcsname
\fi
#2}}
\providecommand{\BIBdecl}{\relax}
\BIBdecl

\bibitem{dantzig1959truck}
G.~B. Dantzig and J.~H. Ramser, ``The truck dispatching problem,'' \emph{Management science}, vol.~6, no.~1, pp. 80--91, 1959.

\bibitem{laporte2007you}
G.~Laporte, ``What you should know about the vehicle routing problem,'' \emph{Naval Research Logistics (NRL)}, vol.~54, no.~8, pp. 811--819, 2007.

\bibitem{arnold2019efficiently}
F.~Arnold, M.~Gendreau, and K.~S{\"o}rensen, ``Efficiently solving very large-scale routing problems,'' \emph{Computers \& operations research}, vol. 107, pp. 32--42, 2019.

\bibitem{gendreau2010solving}
M.~Gendreau and C.~D. Tarantilis, \emph{Solving large-scale vehicle routing problems with time windows: The state-of-the-art}.\hskip 1em plus 0.5em minus 0.4em\relax Cirrelt Montreal, 2010.

\bibitem{sabar2015math}
N.~R. Sabar, X.~J. Zhang, and A.~Song, ``A math-hyper-heuristic approach for large-scale vehicle routing problems with time windows,'' in \emph{2015 IEEE congress on evolutionary computation (CEC)}.\hskip 1em plus 0.5em minus 0.4em\relax IEEE, 2015, pp. 830--837.

\bibitem{costa2020adaptive}
J.~G.~C. Costa, Y.~Mei, and M.~Zhang, ``Adaptive search space through evolutionary hyper-heuristics for the large-scale vehicle routing problem,'' in \emph{2020 IEEE Symposium Series on Computational Intelligence (SSCI)}.\hskip 1em plus 0.5em minus 0.4em\relax IEEE, 2020, pp. 2415--2422.

\bibitem{blum2003metaheuristics}
C.~Blum and A.~Roli, ``Metaheuristics in combinatorial optimization: Overview and conceptual comparison,'' \emph{ACM computing surveys (CSUR)}, vol.~35, no.~3, pp. 268--308, 2003.

\bibitem{wang2021survey}
X.~Wang, Y.~Chen, and W.~Zhu, ``A survey on curriculum learning,'' \emph{IEEE transactions on pattern analysis and machine intelligence}, vol.~44, no.~9, pp. 4555--4576, 2021.

\bibitem{bengio2009learning}
Y.~Bengio, ``Learning deep architectures for {AI},'' 2009.

\bibitem{laporte2009fifty}
G.~Laporte, ``Fifty years of vehicle routing,'' \emph{Transportation science}, vol.~43, no.~4, pp. 408--416, 2009.

\bibitem{jin2018data}
Y.~Jin, H.~Wang, T.~Chugh, D.~Guo, and K.~Miettinen, ``Data-driven evolutionary optimization: An overview and case studies,'' \emph{IEEE Transactions on Evolutionary Computation}, vol.~23, no.~3, pp. 442--458, 2018.

\bibitem{soviany2022curriculum}
P.~Soviany, R.~T. Ionescu, P.~Rota, and N.~Sebe, ``Curriculum learning: A survey,'' \emph{International Journal of Computer Vision}, vol. 130, no.~6, pp. 1526--1565, 2022.

\bibitem{liu2024gpgls}
S.~Liu, J.~G. Cavalcanti~Costa, Y.~Mei, and M.~Zhang, ``Gpgls: Genetic programming guided local search for large-scale vehicle routing problems,'' in \emph{International Conference on Parallel Problem Solving from Nature}.\hskip 1em plus 0.5em minus 0.4em\relax Springer, 2024, pp. 36--51.

\bibitem{queiroga202110}
E.~Queiroga, R.~Sadykov, E.~Uchoa, and T.~Vidal, ``10,000 optimal cvrp solutions for testing machine learning based heuristics,'' in \emph{AAAI-22 workshop on machine learning for operations research (ML4OR)}, 2021.

\bibitem{uchoa2017new}
E.~Uchoa, D.~Pecin, A.~Pessoa, M.~Poggi, T.~Vidal, and A.~Subramanian, ``New benchmark instances for the capacitated vehicle routing problem,'' \emph{European Journal of Operational Research}, vol. 257, no.~3, pp. 845--858, 2017.

\bibitem{arnold2019knowledge}
F.~Arnold and K.~S{\"o}rensen, ``Knowledge-guided local search for the vehicle routing problem,'' \emph{Computers \& Operations Research}, vol. 105, pp. 32--46, 2019.

\bibitem{pei2025adaptive}
J.~Pei, Y.~Mei, J.~Liu, M.~Zhang, and X.~Yao, ``Adaptive operator selection for meta-heuristics: A survey,'' \emph{IEEE Transactions on Artificial Intelligence}, 2025.

\bibitem{pei2022investigation}
J.~Pei, Y.~Mei, J.~Liu, and X.~Yao, ``An investigation of adaptive operator selection in solving complex vehicle routing problem,'' in \emph{Pacific Rim International Conference on Artificial Intelligence}.\hskip 1em plus 0.5em minus 0.4em\relax Springer, 2022, pp. 562--573.

\end{thebibliography}
